\documentclass[journal,12pt,onecolumn,draftclsnofoot]{IEEEtran}

\ifCLASSINFOpdf

\else

\fi
\usepackage{amsmath,epsfig,amssymb,verbatim}
\usepackage{cite}
\usepackage[caption=false]{subfig}
\usepackage{array,algorithm,algorithmic}
\usepackage{comment}
\def\BibTeX{{\rm B\kern-.05em{\sc i\kern-.025em b}\kern-.08em
    T\kern-.1667em\lower.7ex\hbox{E}\kern-.125emX}}

\usepackage{enumitem}
\setlist{leftmargin=5.5mm}
\usepackage{enumitem}
\setlist{leftmargin=4mm}
\usepackage{multirow}
\usepackage{multicol}
\usepackage{}
\usepackage{color}

\begin{document}
\title{An Unsupervised Learning Approach for Spectrum Allocation in Terahertz Communication Systems}

\author{Akram~Shafie, Chunhui~Li,~Nan~Yang, Xiangyun Zhou, and Trung Q. Duong
\thanks{Akram Shafie, Nan Yang, and Xiangyun Zhou are with the School of Engineering, The Australian National University, Canberra, ACT 2601, Australia (e-mail: akram.shafie@anu.edu.au, nan.yang@anu.edu.au).}
\thanks{Chunhui Li is with Ericsson Research, Japan (e-mail: chunhui.li@ericsson.com). }
\thanks{Trung Q. Duong is with the School of Electronics, Electrical Engineering, and Computer Science, Queen's University Belfast, U.K. (e-mail: trung.q.duong@qub.ac.uk). }
}


\maketitle

\begin{abstract}
We propose a new spectrum allocation strategy, aided by unsupervised learning, for multiuser terahertz communication systems. In this strategy, adaptive sub-band bandwidth is considered such that the spectrum of interest can be divided into sub-bands with unequal bandwidths. This strategy reduces the variation in molecular absorption loss among the users, leading to the improved data rate performance. We first formulate an optimization problem to determine the optimal sub-band bandwidth and transmit power, and then propose the unsupervised learning-based approach to obtaining the near-optimal solution to this problem. In the proposed approach, we first train a deep neural network (DNN) while utilizing a loss function that is inspired by the Lagrangian of the formulated problem. Then using the trained DNN, we approximate the near-optimal solutions. Numerical results demonstrate that comparing to existing approaches, our proposed unsupervised learning-based approach achieves a higher data rate, especially when the molecular absorption coefficient within the spectrum of interest varies in a highly non-linear manner.
\end{abstract}

\begin{IEEEkeywords}
Terahertz communication, machine learning, unsupervised learning, spectrum allocation, adaptive bandwidth
\end{IEEEkeywords}

\section{Introduction}
Terahertz (THz) communication (THzCom) has been envisioned as a key wireless technology in the sixth-generation (6G) and beyond era \cite{2020_Mag6G_Marco_UseCasesandTechnologies}. The THz band has enormous potential to support tera-bits-per-second (Tbps) data rates and massive connections in 6G networks, due to the tens up to a hundred GHz bandwidth~\cite{akram2022IEEENetwork}. Thanks to the recent advancements in manufacturing THz transceivers and antennas, as well as the experimental licenses (95 GHz--3 THz) opened by the US Federal Communications Commission, increasing endeavors have been devoted to the design and development of practical THzCom systems over the past few years \cite{2022_Sarieddeen_ComMag}.

Despite the promise, THzCom encounters unique challenges that are different from those experienced at lower frequencies, e.g., severe spreading loss, higher channel sparsity, and the unique molecular absorption loss \cite{2020_WCM_THzMag_TerahertzNetworks}.
The molecular absorption loss is frequency-dependent, divides the whole THz band into several ultra-wide transmission windows, and introduces substantially varying distance-dependent path loss within transmission windows. These challenges need to be wisely tackled for developing practical THzCom systems.

There have recently been increasing endeavors to exploring novel and efficient THz spectrum allocation strategies, e.g., \cite{2020_WCM_THzMag_TerahertzNetworks,2020_WCM_THzMag_Standardization,akram2022IEEENetwork}, due to its benefits of exploiting the potentials of the THz band. Specifically, multi-band-based spectrum allocation is envisioned as one promising solution to supporting micro- and macro-scale THzCom systems, since it can effectively assign spectral resources when the variation in molecular absorption loss among the users in multiuser systems is very high \cite{HBM2}. In this allocation, the spectrum of interest is divided into non-overlapping sub-bands, and the sub-bands are utilized to support the users in the system.

Although different designs have been proposed to optimize the performance of multi-band-based spectrum allocation, e.g., \cite{HBM2,2017N5,HBM3,2019_Chong_DABM2}, they all considered the spectrum of interest is divided into sub-bands with equal bandwidth. It is noted that the consideration of equal sub-band bandwidth (ESB) can lead to high variation in the molecular absorption loss among the sub-bands. Different from ESB, adaptive sub-band bandwidth (ASB), where the spectrum of interest is divided into sub-bands with unequal bandwidths, can effectively reduce the variation in molecular absorption loss among the users, thus leading to an overall improvement in the data rate performance.
Motivated by this, the multi-band-based spectrum allocation with ASB was first proposed in our previous studies \cite{akram2021TCOM,akram2022TVT}. We note that the design in \cite{akram2021TCOM} is only applicable when the molecular absorption coefficient within the to-be-allocated spectrum is simple such that it can be modeled as a piecewise exponential function of frequency with minimal approximation errors. In many spectrum regions within the THz band, the molecular absorption coefficient varies in a highly non-linear manner; thus, it cannot be modeled as a piecewise exponential function of frequencies in such spectrum regions. It follows that the designs in \cite{akram2021TCOM,akram2022TVT} cannot be adopted in such spectrum regions. This motivates the current work.

In the current work, we study multi-band-based spectrum allocation with ASB to improve the spectral efficiency of THzCom systems. We  formulate an optimization problem  to determine the optimal sub-band bandwidth and optimal transmit power.
We then propose an unsupervised learning-based approach to obtaining a near-optimal solution to the formulated optimization problem. In the proposed approach, we first train a deep neural network (DNN) where a loss function inspired by the Lagrangian of the formulated optimization problem is used. Thereafter, using the trained DNN, we approximate the near-optimal solution to the formulated optimization problem. Using numerical results, we demonstrate that when the values of the molecular absorption coefficient within the spectrum of interest can be modeled as an exponential function of frequency, the data rate obtained by our proposed unsupervised learning-based approach converges to the optimal data rate achieved by the existing spectrum allocation approach with ASB. We also demonstrate that when the values of the molecular absorption coefficient within the spectrum of interest change rapidly and thus cannot be modeled as an exponential function of frequency, our proposed unsupervised learning-based approach outperforms the existing approach.



\textit{Notations:} Matrices and column vectors are denoted by uppercase and lowercase boldface symbols, respectively. Scalar variables are denoted by italic symbols. Given a matrix $\mathbf{A}$, $\mathbf{A}_{\mathrm{r}}$ denotes the $r$th row vector of $\mathbf{A}$. Given a column vector $\mathbf{a}$, $\mathbf{a}^{\mathrm{T}}$ and $a_{\mathrm{r}}$ denote the transpose and the $r$th element of $\mathbf{a}$, respectively. Moreover, $\mathbf{I}_m$ and $\mathbf{L}_{m}$ denote $m\times m$ identity matrix and $m\times m$ lower triangular matrix with all non-zero values as 1, respectively. Furthermore, $\mathbf{1}_{m}$ denotes the $m\times 1$ vector with all entries as 1.
The curled inequality symbols, e.g., $\preceq$,  denote generalized inequalities, i.e., componentwise inequality between vectors.

\section{System Model and Problem Formulation}\label{Sec:System}





We consider a  three-dimensional (3D) indoor THzCom system where the high data rate demands of $n_{\mathrm{I}}$ users  are supported by  a single AP. 
We consider that the users are stationary and distributed uniformly on the floor in the indoor environment.
We denote $\mathbf{d}$ as the $n_{\mathrm{I}}\times 1$ communication distance vector of the AP-user channels.
For notational convenience, we consider that the elements in $\mathbf{d}$  are ordered such that $d_1< d_2< \cdots < d_{n_{\mathrm{I}}}$.


\subsection{THz Spectrum}

\subsubsection{Spectrum of Interest}

\begin{figure}[!t]
\centering
\vspace{-4mm}
\includegraphics[width=0.7\columnwidth]{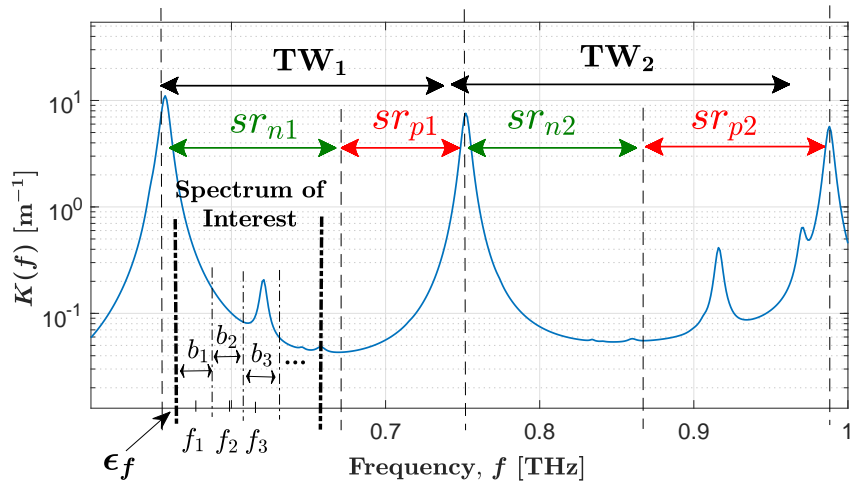}
\vspace{-4mm}
\caption{Illustration of (i) PACSRs  and NACSRs that exist between 0.5 THz and 1 THz  with $\{\mathrm{TW}_{1}, \mathrm{TW}_{2}\}$, $\{sr_{p1},sr_{p2}\}$, and  $\{sr_{n1},sr_{n2}\}$ denoting the transmission windows, PACSRs, and NACSRs, respectively, and (ii) sub-band arrangement within the spectrum of interest.}\label{Fig:THzBand} \vspace{-2mm}
\end{figure}

We define the regions which generally exhibit an decreasing and increasing molecular absorption coefficient behaviour within an ultra-wideband THz transmission window as the negative absorption coefficient slope region (NACSR) and the positive absorption coefficient slope region (PACSR), respectively, as depicted in  Fig.~\ref{Fig:THzBand} \cite{akram2021TCOM}. It is noted that at the THz band, the available bandwidths in each NACSR and PACSR are in the order of  tens of GHz \cite[Table~1]{akram2021TCOM}.
Considering this, we focus on the scenario where the to-be-allocated spectrum of interest fully exists in either an NACSR or a PACSR of a THz transmission window in this work\footnote{To fully utilize the potentials of the huge available bandwidths at the THz band, it would be more beneficial to focus on resource allocation when the to-be-allocated spectrum of interest spans across multiple NACSRs and/or PACSRs. This consideration will be addressed in our future work.}. Without loss of generality, in the rest of the paper, we present spectrum allocation in an NACSR only since this spectrum allocation can be easily applied to a PACSR.

We focus on multi-band-based spectrum allocation with ASB. Thus, we divide the spectrum of interest into $n_{\mathrm{S}}$ sub-bands that are of unequal bandwidths. We denote $\mathbf{b}$ and $\mathbf{f}$ as the $n_{\mathrm{S}}\times 1$ vectors of the bandwidth and the center frequency of the sub-bands, respectively.
For notational convenience, we label the sub-bands such that $f_1< f_2< \cdots < f_{n_{\mathrm{S}}}$, as shown in Fig.~\ref{Fig:THzBand}.
Thus, we have \vspace{-2mm}
\begin{align}
f_{s}=\epsilon_f + \sum\nolimits_{k=1}^{s-1}b_{k}+\frac{1}{2}b_{s}= \mathbf{A}_s\mathbf{b}+\epsilon_f,
\end{align}
where $s\in\{1,2,\cdots,n_{\mathrm{S}}\}$ and $\epsilon_f$  is the start frequency of the spectrum of interest, as shown in Fig.~\ref{Fig:THzBand}. Here, $\mathbf{A}$ is an $n_{\mathrm{S}}\times n_{\mathrm{S}}$ matrix with $\mathbf{A}=\mathbf{L}_{n_{\mathrm{S}}}-\frac{1}{2}\mathbf{I}_{n_{\mathrm{S}}}$.
Considering this, we have $ \mathbf{f}= \mathbf{A}\mathbf{b}+\mathbf{1}_{n_{\mathrm{I}}}\epsilon_f$.
Due to the consideration of ASB, we have  \vspace{-1mm}
\begin{align}\label{Equ:BmaxConst}
0 \preceq \mathbf{b} \preceq b_{\textrm{max}}, \end{align}
where $b_{\textrm{max}}$  denotes the upper bound on bandwidth of the sub-band. 
We then denote $b_{\textrm{tot}}$ as the total available bandwidth within the spectrum of interest and express it as\vspace{-1mm}
\begin{align}\label{Equ:BtotConst}
 b_{\textrm{tot}}=\mathbf{1}_{n_{\mathrm{S}}}^{\textrm{T}} \mathbf{b}.
\end{align}

\subsubsection{Sub-band Assignment}

In the considered system, we assume that the users in the system are served by separate sub-bands\footnote{A higher spectral efficiency can be achieved by utilizing sub-band reuse during spectrum allocation, which will be considered in our
future work.}. This assumption is made to (i) ensure the data transmission to be free of intra-band interference and (i) eliminate the signal processing overhead and hardware complexity caused by frequency reuse in the system, which is also adopted in  \cite{2017N5,HBM2,HBM3,akram2021TCOM,akram2022TVT}. Considering this assumption, we set the total number of sub-bands within the spectrum of interest to be equal to the total number of users in the system, i.e., $n_{\mathrm{S}}=n_{\mathrm{I}}$.
Also, following \cite{HBM3,HBM2,2019_Chong_DABM2}, we adopt  distance-aware multi-carrier (DAMC)-based sub-band assignment to improve the throughput fairness among users.
The DAMC-based sub-band assignment assigns (i) the sub-bands with high absorption coefficients to the users with shorter distances and (ii) the sub-bands with low absorption coefficients to the users with longer distances.


\subsection{Achievable Data Rate}

\label{Sec:Channel}

We denote $\mathbf{r}$ as the $n_{\mathrm{S}}\times 1$ rate vector of users. Considering spreading and molecular absorption losses experienced by THz signal propagation, we obtain the rate achieved in the $s$th sub-band as \cite{2011_Jornet_TWC}
\begin{align} \label{Equ:Rateijs1}
&r_{s}=\int\nolimits_{f_s-\frac{1}{2}b_s}^{f_s+\frac{1}{2}b_s}  \log_{2}\left(1+\frac{p_{s} \varrho e^{-k(f) d_s}}{f^2 d^2_s b_{s}}\right) \textrm{d}f ,
\end{align}
where $\varrho\;\triangleq G_{\textrm{A}}G_{\textrm{U}}N_{0}^{-1}\left(\frac{c}{4 \pi }\right)^{2}$, $G_{\textrm{A}}$ and $G_{\textrm{U}}$ are the antenna gains at the AP and users, respectively, $N_{0}$ is the noise power density, $c$ is the speed of light, and $k(f)$ is the molecular absorption coefficient at $f$. Furthermore, $\mathbf{p}$ denotes the $n_{\mathrm{S}}\times 1$ transmit power vector.

In this work, we focus on the line-of-sight (LoS) rays of THz signals  \cite{2011_Jornet_TWC,akram2020JSAC,akram2021TCOM,akram2022TVT}. This is reasonable because when THz signals propagate, the direct ray dominates the received signal energy, while the non-line-of-sight  rays are significantly attenuated due to high scattering and reflection losses.  Also, we omit the impact of fading in the THz channel as the prior studies at the THz band~\cite{2011_Jornet_TWC,akram2020JSAC,akram2021TCOM,akram2022TVT}.
Moreover, following \cite{HBM2,2017N5,HBM3,2019_Chong_DABM2,akram2022TVT}, we omit the impact of blockages in the THz channel.
Finally, the impact of inter-band interference (IBI) is not considered in this work since prior studies have proposed designs to suppress IBI with minimal throughput degradation~\cite{2020_Chong_TWC_DistanceAdaptiveAbsorptionPeakModulation}.

\subsection{Optimal Spectrum Allocation}\label{Sec:ProbForm}


We now design an  efficient spectrum allocation strategy to harness the potential of the THz band. To this end, we study  spectrum allocation  with ASB to maximize the achievable data rate under given sub-band bandwidth and power constraints.
Mathematically, this problem is formulated as \vspace{-1mm}
\begin{subequations}\label{OptProbOrg}
\begin{alignat}{2}
 \mathbf {P^{o}:}& \quad  \underset{\substack{\mathbf{p}, \mathbf{b}}}{\textrm{max}}
& & \quad \mathcal {E}(\mathbf{d},\mathbf{p}, \mathbf{b})                                     \label{OptProbOrgA} \\
&\quad    \textup{s. t.}
 & &  \quad \mathbf{1}_{n_{\mathrm{S}}}^{\textrm{T}}\mathbf{p} \leqslant p_{\textrm{tot}},     \label{OptProbOrgB}\\
& & &  \quad 0 \preceq \mathbf{p} \preceq p_{\textrm{max}},                                     \label{OptProbOrgC} \\
& & &  \quad \mathbf{1}_{n_{\mathrm{S}}}^{\textrm{T}}\mathbf{b} = b_{\textrm{tot}},      \label{OptProbOrgD} \\
& & &  \quad 0 \preceq \mathbf{b} \preceq b_{\textrm{max}}.                        \label{OptProbOrgE}
\end{alignat}
\end{subequations}

\noindent In $\mathbf {P^{o}}$, $\mathcal {E}(\mathbf{d},\mathbf{p}, \mathbf{b})$ is the objective function of the considered data rate maximization strategy. In this work, we consider proportionally-fair data rate maximization, which leads to $\mathcal {E}(\mathbf{d},\mathbf{p}, \mathbf{b}) = \mathbf{1}^{\textrm{T}}_{n_{\mathrm{S}}} \log (\mathbf{r})$.
Moreover, \eqref{OptProbOrgB} and \eqref{OptProbOrgC} are the power budget at the AP and the upper bound on the power allocated to each user, respectively.
Furthermore, the justifications on \eqref{OptProbOrgD} and \eqref{OptProbOrgE} are given by \eqref{Equ:BtotConst} and \eqref{Equ:BmaxConst}, respectively.

We note that it is extremely difficult, if not impossible, to analytically solve $\mathbf {P^{o}}$  using the traditional optimization techniques \cite{ConvexBoyedBook}. This is due to the
difficulty in obtaining a tractable expression for $r_s$ in terms of the design variables $\mathbf{b}$. On one hand, obtaining $r_s$ as per \eqref{Equ:Rateijs1} involves an integral, the limits of which depend on $\mathbf{b}$. On the other hand, the values of $k(f)$ for all frequencies within the spectrum of interest are required to obtain $r_s$. However, there does not exist a tractable expression that maps $f$ to $k(f)$ for all spectrum regions within the THz band. To tackle these challenges,  we resort to unsupervised learning to obtain near-optimal solution to $\mathbf {P^{o}}$ \cite{2019_TSP_DNN_Mark,2022_Chunhui_MLforSecurity}, and present the solution in Section \ref{Sec:UL_Solution}.

\section{Unsupervised Learning-Based Solution}\label{Sec:UL_Solution}

In this section, we present an unsupervised learning-based approach to solving problem $\mathbf {P^{o}}$ in \eqref{Equ:Rateijs1}.
In this approach, we first employ an \textit{offline training phase} to train a DNN, utilizing (i) a loss function which is inspired by the objective function of the dual problem of $\mathbf {P^{o}}$ and (ii) a batch of realizations of distance vectors. Thereafter, during the \textit{implementation phase}, we use the trained DNN to approximate the optimal solution to $\mathbf {P^{o}}$ for the given distance vector.
Next, we will discuss the architecture of the adopted DNN in Section \ref{Sec:DNNArchitecture}, and the unsupervised learning model used in the training phase in Section \ref{Sec:DLModel}.


\subsection{Architecture of DNN}\label{Sec:DNNArchitecture}

\begin{figure}[!t]
\centering
\includegraphics[width=0.7\columnwidth]{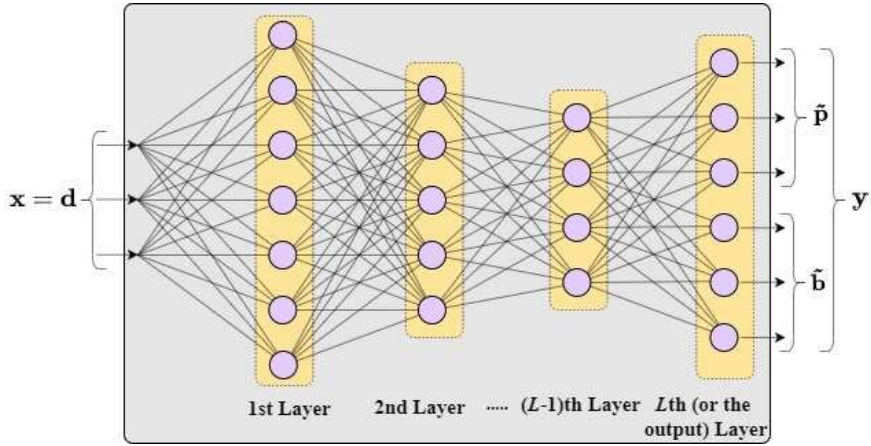}
\vspace{-5mm}
\caption{Illustration of the general architecture of the DNN when $L=4$ and $n_{\mathrm{S}}=3$.}\label{Fig:DNN} \vspace{-2mm}
\end{figure}

\begin{figure}[t]
\centering
\includegraphics[width=0.995\columnwidth]{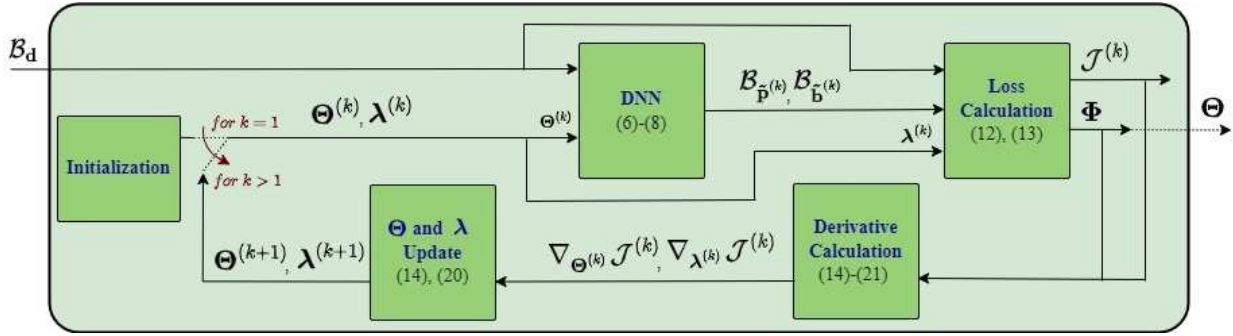}
\vspace{-1mm}
\caption{Block diagram representation of the unsupervised learning model. Here, $\mathcal{B}_{\mathbf{\tilde{p}
}^{(k)}}$ and $\mathcal{B}_{\mathbf{\tilde{b}
}^{(k)}}$ are the batches of $\mathbf{\tilde{p}}$ and $\mathbf{\tilde{b}}$ obtained as the output of DNN at the $k$th iteration, and $\boldsymbol \Phi=\{\mathcal{B}_{\mathbf{\tilde{p}
}^{(k)}},\mathcal{B}_{\mathbf{\tilde{b}
}^{(k)}},\boldsymbol \Theta^{(k)},\boldsymbol \lambda^{(k)}\}$ denotes  the cache of the unsupervised learning  model.}\label{Fig:DLModel} \vspace{-2mm}
\end{figure}

The general architecture of the adopted DNN is presented in Fig.~\ref{Fig:DNN}.
The DNN consists of $L$ fully-connected layers with multiple neurons in every layer.
Each neuron in the network contains a linear operation, which is followed by a point-wise nonlinearity, also known as an activation function.
Considering this, the output of the network, $\mathbf{y}$, is expressed as a non-linear function to the input  of the network, $\mathbf{x}$, which is given by
\begin{equation}
\mathbf{y} = {{\mathcal{N}}}({\mathbf{x},{\boldsymbol \Theta}}) = {g^{(L)}}\left( {{g^{(L - 1)}}\left( { \cdots{g^{(1)}}\left(\mathbf{x}\right)} \right)} \right).
\end{equation}
Here, 
${\boldsymbol \Theta}=\{{\boldsymbol \varpi}^{(\ell)},\beta^{(\ell)}, \ell\in\{1,2,\cdots,L\}\}$, denotes the set of parameters (i.e., weights and biases) of the DNN, and ${\boldsymbol \varpi}^{(\ell)}$ and $\beta^{(\ell)}$ are the weights vector and the bias of the $\ell$th layer, respectively.
Also, we have
\begin{equation}{g^{(\ell)}} (\boldsymbol \rho)=\sigma^{(\ell)} \left(({\boldsymbol \varpi}^{(\ell)})^\mathrm{T}{{\boldsymbol \rho}} + {\beta^{(\ell)}}\right),\end{equation}
where $\sigma^{(\ell)} (\cdot)$ is the activation function  of the $\ell$th layer.

During the \textit{implementation phase}, we provide $\mathbf{d}$ as the input to the DNN and obtain the output as the approximated values of the optimization variables of $\mathbf {P^{o}}$, which are denoted by $\mathbf{\tilde{p}}$ and $\mathbf{\tilde{b}}$. Therefore, we have \vspace{-3mm}
\begin{align}
\mathbf{x}=\mathbf{d} ~~~~~ \mathrm{and}~~~~
\mathbf{y} &= \begin{bmatrix}
\mathbf{\tilde{p}}\\
\mathbf{\tilde{b}}
\end{bmatrix}. \end{align}


\subsection{Unsupervised Learning Model}\label{Sec:DLModel}

The block diagram representation of the unsupervised learning model used in the training phase is depicted in Fig.~\ref{Fig:DLModel}. The unsupervised learning model takes a batch of realization of the distance vector, $\mathcal{B}_{\textbf{d}}$, as the input, and delivers the trained ${\boldsymbol \Theta}$ as the output.
We note that the unsupervised learning  model iteratively updates ${\boldsymbol \Theta}$ to ensure the best mapping being obtained between $\mathcal{B}_{\textbf{d}}$ and the near-optimal solution to $\mathbf {P^{o}}$ for the given $\mathcal{B}_{\textbf{d}}$.

\subsubsection{Loss Function}

The loss function of the unsupervised learning model is inspired by the objective function of the dual problem of $\mathbf {P^{o}}$. Specifically, the dual problem of $\mathbf {P^{o}}$ is given by
\begin{subequations}\label{OptProbOrg2}
\begin{alignat}{2}
\mathbf {P_{\!D}^{o}:} ~~& \underset{\substack{\boldsymbol \lambda, \boldsymbol \Gamma}}{\textrm{max}} ~~ \underset{\substack{\mathbf{p},\mathbf{b}}}{\textrm{min}} \  \mathcal{L}_{\mathrm{o}}
\\ &  \mathrm{s}.\mathrm{t}.~~~ \eqref{OptProbOrgB}-\eqref{OptProbOrgE} \notag \\ &  ~~~~~~\ \boldsymbol \lambda, \boldsymbol \Gamma\succeq 0,
\end{alignat}
\end{subequations}
where
\begin{align} \label{Equ:Lorg} &\mathcal{L}_{\mathrm{o}}=\mathcal{L}_{\mathrm{o}}(\mathbf{d},\mathbf{p}, \mathbf{b},\boldsymbol \lambda, \boldsymbol \Gamma)\notag\\
&=-\mathcal {E}(\mathbf{d},\mathbf{p}, \mathbf{b})+ \lambda_1\left(\mathbf{1}^{\textrm{T}}_{n_{\mathrm{S}}}\mathbf{p}- p_{\textrm{tot}}\right)+ \lambda_2\left(\mathbf{1}^{\textrm{T}}_{n_{\mathrm{S}}}\mathbf{b}- b_{\textrm{tot}}\right)\notag\\
&~~~{+}\boldsymbol \Gamma^{\textrm{T}}_1 (0{-}\mathbf{p}){+}\boldsymbol \Gamma^{\textrm{T}}_2 (\mathbf{p}{-}p_{\textrm{max}}){+}\boldsymbol \Gamma^{\textrm{T}}_3 (0{-}\mathbf{b}){+}\boldsymbol \Gamma^{\textrm{T}}_4 (\mathbf{b}{-}b_{\textrm{max}}) \end{align}
is the Lagrangian of $\mathbf {P^{o}}$, and $\boldsymbol\lambda=[\lambda_1, \lambda_2]^{\mathrm{T}}$ and $\boldsymbol\Gamma=[\boldsymbol {\Gamma_1, \Gamma_2, \Gamma_3,\Gamma_4}]$ are the Lagrange multipliers associated with the constraints of $ \mathbf {P^{o}}$.
In our unsupervised learning model, $\mathbf{p}$ and $\mathbf{b}$ are approximated as $\tilde{\mathbf{p}}$ and $\tilde{\mathbf{b}}$ using the DNN, respectively, which is parameterized by ${\boldsymbol \Theta}$.
Considering this, the problem $\mathbf {P_{\!D}^{o}}$ can be regenerated into the following variable optimization as
\begin{subequations}\label{OptProbOrg3}
\begin{alignat}{2}
&\mathbf {\hat{P}^{o}}:~~ & \underset{\substack{\boldsymbol \lambda}}{\textrm{max}} &~~\underset{\substack{{\boldsymbol \Theta}}}{\textrm{min}} \  \hat{\mathcal{L}}
\\ &&  \mathrm{s}.\mathrm{t}. &~~\eqref{OptProbOrgB},\eqref{OptProbOrgD} \notag \\ &&  &~~ \boldsymbol \lambda\succeq 0,
\end{alignat}
\end{subequations}
where
\begin{align} \label{Equ:Lmod}  \hat{\mathcal{L}}&=\hat{\mathcal{L}}(\mathbf{d},\tilde{\mathbf{p}}, \tilde{\mathbf{b}},{\boldsymbol \lambda})\notag\\
&=-\mathcal {E}(\mathbf{d},\tilde{\mathbf{p}}, \tilde{\mathbf{b}}){+}\lambda_1\left(\mathbf{1}^{\textrm{T}}_{n_{\mathrm{S}}}\tilde{\mathbf{p}}{-} p_{\textrm{tot}}\right){+} \lambda_2\left(\mathbf{1}^{\textrm{T}}_{n_{\mathrm{S}}}\tilde{\mathbf{b}}{-} b_{\textrm{tot}}\right).\end{align}
In \eqref{Equ:Lmod}, $\hat{\mathcal{L}}$ is obtained as a simplified version of $\mathcal{L}_{\mathrm{o}}$ given in \eqref{Equ:Lorg}. This is because the satisfaction of constraints \eqref{OptProbOrgC} and \eqref{OptProbOrgE} can be guaranteed by carefully choosing appropriate activation functions for the neurons in the $L$th (or the output) layer of the DNN, e.g., $\sigma^{(L)}(\cdot)=\vartheta\times S(\cdot)$, where $S(\cdot)$ is the sigmoid function,  and $\vartheta=p_{\textrm{max}}$ and $\vartheta=b_{\textrm{max}}$ for the first and the second $n_{\mathrm{S}}$ neurons of the output layer, respectively.

We note that the optimal value of $\mathbf {{P}_{D}^{o}}$ might be a loose global lower bound on the optimal value of $\mathbf {{P}^{o}}$, since $\mathbf{{P}^{o}}$ is not a convex optimization problem.
However, we also note that the design variables of $\mathbf {{P}_{D}^{o}}$ are approximated using a DNN in our unsupervised learning model and $\mathbf {\hat{P}^{o}}$ aims to optimize the DNN parameters, ${\boldsymbol \Theta}$.
Thus, it can be concluded based on the universal approximation theorem that the optimal value of $\mathbf {{\hat{P}}^{o}}$ provides a tight lower bound on the optimal value of $\mathbf {{P}^{o}}$ \cite{2019_TSP_DNN_Mark,2022_Chunhui_MLforSecurity}.
Based on this conclusion and considering that a batch of realization of $\textbf{d}$, $\mathcal{B}_{\textbf{d}}$, is used during the training phase,  we define the loss function of the unsupervised learning model, $\mathcal{J}$. Specifically, for each realization of $\textbf{d}$ in $\mathcal{B}_{\textbf{d}}$, we obtained the corresponding $\hat{\mathcal{L}}$, and then derive $\mathcal{J}$ as the mean of $\hat{\mathcal{L}}$ values. Mathematically, $\mathcal{J}$ is obtained as
\begin{align}\label{A}
\mathcal{J}=\mathcal{J}(\boldsymbol\Theta, \boldsymbol \lambda)=\frac{1}{n_{\mathrm{T}}} \sum\nolimits_{\mathbf{d}\in\mathcal{B}_{\textbf{d}}}\hat{\mathcal{L}},
\end{align}
where $n_{\mathrm{T}}$ is the total number of realization of $\textbf{d}$ in $\mathcal{B}_{\textbf{d}}$. 
We then iteratively update ${\boldsymbol\Theta}$ and $\boldsymbol \lambda$  with the objective of minimizing $\mathcal{J}$ to arrive at the near-optimal solution to $\mathbf {{P}^{o}}$.


\subsubsection{Updating ${\boldsymbol\Theta}$ and $\boldsymbol \lambda$}

The DNN parameters, ${\boldsymbol \Theta}$,  are updated according to
\begin{align}  \boldsymbol \Theta^{(k+1)}&= \boldsymbol \Theta^{(k)}-\delta_{ \Theta} ~\nabla_{\boldsymbol\Theta^{(k)}}\mathcal{J}^{(k)}\notag\\
&=\boldsymbol \Theta^{(k)}-\frac{\delta_{ \Theta}}{n_{\mathrm{T}} } \sum_{\mathbf{d}\in \mathcal{B}_{\textbf{d}}} \left[\nabla_{\boldsymbol\Theta}\hat{\mathcal{L}}\right]_{\substack{\boldsymbol\Theta=\boldsymbol\Theta^{(k)},\tilde{\mathbf{p}}=\tilde{\mathbf{p}}^{(k)},\\\tilde{\mathbf{b}}=\tilde{\mathbf{b}}^{(k)}, {\boldsymbol \lambda}={\boldsymbol \lambda}^{(k)}}}
\label{Equ:Theta_update}\end{align}
where $\delta_{ \Theta}$ is the learning rate and ${\boldsymbol\Psi}^{(k)}$ is the $k$th iteration value of ${\boldsymbol\Psi}$, with ${\boldsymbol\Psi}\in\{\boldsymbol \Theta,\mathcal{J}, \tilde{\mathbf{p}}, \tilde{\mathbf{b}},{\boldsymbol \lambda}\}$.
Considering the chain rule, we express $\nabla_{\boldsymbol\Theta}\hat{\mathcal{L}}$ in \eqref{Equ:Theta_update} as
\begin{align} &\nabla_{\boldsymbol\Theta}\hat{\mathcal{L}}= \nabla_{\boldsymbol\Theta}\mathbf{y}\begin{bmatrix}
\nabla_{\tilde{\mathbf{p}}}\hat{\mathcal{L}}\\ \nabla_{\tilde{\mathbf{b}}}\hat{\mathcal{L}}
\end{bmatrix} \label{Equ:dLdTheta}
\end{align}
In \eqref{Equ:dLdTheta}, $\nabla_{\tilde{\mathbf{p}}}\hat{\mathcal{L}}$ and  $\nabla_{\tilde{\mathbf{b}}}\hat{\mathcal{L}}$ can be derived as
\begin{align} \nabla_{\tilde{\mathbf{p}}}\hat{\mathcal{L}}&= -\nabla_{\tilde{\mathbf{p}}}\mathcal {E}(\mathbf{d},{\tilde{\mathbf{p}}}, {\tilde{\mathbf{b}}})+\lambda_1, \label{Equ:dLdp}
\end{align}
and
\begin{align} \nabla_{\tilde{\mathbf{b}}}\hat{\mathcal{L}}&=- \nabla_{\tilde{\mathbf{b}}}\mathcal {E}(\mathbf{d},{\tilde{\mathbf{p}}}, {\tilde{\mathbf{b}}})+\lambda_2, \label{Equ:dLdb}
\end{align}
respectively. Note that in \eqref{Equ:dLdp} and \eqref{Equ:dLdb}, it is impossible to analytically derive $\nabla_{\tilde{\mathbf{p}}}\mathcal {E}(\mathbf{d},{\tilde{\mathbf{p}}}, {\tilde{\mathbf{b}}})$ and $\nabla_{\tilde{\mathbf{b}}}\mathcal {E}(\mathbf{d},{\tilde{\mathbf{p}}}, {\tilde{\mathbf{b}}})$, respectively. Hence, we numerically calculate the values of $\nabla_{\tilde{\mathbf{p}}}\mathcal {E}(\mathbf{d},{\tilde{\mathbf{p}}}, {\tilde{\mathbf{b}}})$ and $\nabla_{\tilde{\mathbf{b}}}\mathcal {E}(\mathbf{d},{\tilde{\mathbf{p}}}, {\tilde{\mathbf{b}}})$ when $\tilde{\mathbf{p}}=\tilde{\mathbf{p}}^{(k)}$ and $\tilde{\mathbf{b}}=\tilde{\mathbf{b}}^{(k)}$ by using
\begin{align} \!\!\left[\nabla_{\tilde{\mathbf{p}}}\mathcal {E}(\mathbf{d},\tilde{\mathbf{p}}, \tilde{\mathbf{b}})\right]_{\substack{\tilde{\mathbf{p}}{=}\tilde{\mathbf{p}}^{(k)},\\\\\tilde{\mathbf{b}}{=}\tilde{\mathbf{b}}^{(k)}}}&\!{=}\frac{\mathcal {E}(\mathbf{d},\tilde{\mathbf{p}}^{(k)}{+}\varepsilon\tilde{\mathbf{p}}^{(k)}, \tilde{\mathbf{b}}){-}\mathcal {E}(\mathbf{d},\tilde{\mathbf{p}}^{(k)}, \tilde{\mathbf{b}})}{\varepsilon\tilde{\mathbf{p}}^{(k)}},
\end{align}
and
\begin{align} \!\!\left[\nabla_{\tilde{\mathbf{b}}}\mathcal {E}(\mathbf{d},\tilde{\mathbf{p}}, \tilde{\mathbf{b}})\right]_{\substack{\tilde{\mathbf{p}}{=}\tilde{\mathbf{p}}^{(k)},\\\\\tilde{\mathbf{b}}{=}\tilde{\mathbf{b}}^{(k)}}}&\!{=}\frac{\mathcal {E}(\mathbf{d},\tilde{\mathbf{p}}^{(k)}, \tilde{\mathbf{b}}{+}\varepsilon\tilde{\mathbf{b}}^{(k)}){-}\mathcal {E}(\mathbf{d},\tilde{\mathbf{p}}^{(k)}, \tilde{\mathbf{b}})}{\varepsilon\tilde{\mathbf{b}}^{(k)}},
\end{align}
respectively, where $\varepsilon$ is a very small positive number.
We also note that in \eqref{Equ:dLdTheta}, $\nabla_{\boldsymbol\Theta}\mathbf{y}$ can be calculated as a function of $\boldsymbol \Theta^{(k)}$ utilizing the chain rule.

As for the Lagrangian multiplier, ${\boldsymbol \lambda}$, it can be updated according to
\begin{align} \label{Equ:lambdaUpdate} {\boldsymbol \lambda}^{(k+1)}&{=} \left[\boldsymbol \lambda^{(k)}+\delta_{ \lambda} ~\nabla_{\boldsymbol\lambda^{(k)}}\mathcal{J}^{(k)}\right]^+\notag\\
&{=}\left[\boldsymbol \lambda^{(k)}{+}\frac{\delta_{\lambda}}{n_{\mathrm{T}} }\!\! \sum_{\mathbf{d}\in \mathcal{B}_{\textbf{d}}}\!\!\left[\nabla_{\lambda}\hat{\mathcal{L}}\right]_{\substack{\tilde{\mathbf{p}}=\tilde{\mathbf{p}}^{(k)},\tilde{\mathbf{b}}=\tilde{\mathbf{b}}^{(k)}}}\right]^+\!\!\!\!,\end{align}
where $\delta_{ \lambda}$ is the learning rate. 
In \eqref{Equ:lambdaUpdate}, $\nabla_{\boldsymbol \lambda}\hat{\mathcal{L}}$ can be derived as
\begin{align}
\nabla_{\boldsymbol \lambda}\hat{\mathcal{L}}&= \begin{bmatrix}
\mathbf{1}^{\textrm{T}}_{n_{\mathrm{S}}}\tilde{\mathbf{p}}- p_{\textrm{tot}}\\
\mathbf{1}^{\textrm{T}}_{n_{\mathrm{S}}}\tilde{\mathbf{b}}- b_{\textrm{tot}}
\end{bmatrix}. \end{align}

\noindent
Finally, we note that based on \eqref{Equ:Theta_update} and \eqref{Equ:lambdaUpdate}, we can iteratively update ${\boldsymbol\Theta}$ and $\boldsymbol \lambda$ in the unsupervised learning model.

\section{Convex Optimization-based Solution for a Special Case}\label{Sec:Special_Case}


As mentioned in Section \ref{Sec:UL_Solution}, it is impossible to analytically solve the spectrum allocation problem $\mathbf {P^{o}}$  using traditional optimization techniques for the generalized system model considered in this work. Despite so, for a special case of the considered system model, the solution to $\mathbf {P^{o}}$ can be obtained using the convex optimization theory and standard convex problem solvers~\cite{akram2021TCOM,akram2022TVT}.
We next present this special case system and its solution using convex optimization, the performance of which will be used as a benchmark in Section \ref{Sec:Num} for demonstrating the effectiveness and benefits of our proposed approach in Section \ref{Sec:UL_Solution}.

In the considered special case system, the molecular absorption coefficient within the spectrum of interest, $k(f)$, can be modeled as an exponential function of frequency  with minimal approximation errors.
We note that there indeed exist PACSRs and NACSRs within the THz band where $k(f)$ can be modeled as an exponential function of frequency, e.g., between $0.771$ and $0.821~\textrm{THz}$, where the maximum modeling approximation error is $5\%$. 



For the considered special case system, the data rate, $r_s$, can be approximated as a tractable expression in terms of $\textbf{b}$. Specifically, by modeling ${k}(f)$ within NACSRs as an exponential function of frequency, we can approximate $k({f})$ as
\begin{equation}\label{Equ:CurveFit}
  {k}(f)=e^{\eta_{1} +\eta_{2} f}+\eta_{3},
\end{equation}
where $\boldsymbol \eta=[\eta_1, \eta_2, \eta_3]^{\mathrm{T}}$ is the model parameter vector.
This enables us to obtain $r_{s}$ as a tractable expression of $\textbf{b}$ as
\begin{align} \label{Equ:Rateijs1Mod2}
{r}_{s}&= {b}_{s}\log_{2}\left(1+\frac{{p}_{s}\varrho e^{-{d}_{s}(e^{\eta_{1} +\eta_{2}(\mathbf{A}_s\mathbf{b}+\epsilon_f)}+\eta_{3})}}{ b_{s}d_s^2(\mathbf{A}_s\mathbf{b}+\epsilon_f)^2}\right).
\end{align}



With the simplified expression $r_s$ in~\eqref{Equ:Rateijs1Mod2}, although $\mathbf {P^{o}}$ becomes tractable, it is still non-convex. To overcome the non-convexity, we consider the following substitution for $\mathbf{b}$, given by\vspace{-2mm}
\begin{equation}\label{Equ:Trans1}
  \mathbf{b}=\xi_1+\xi_2\log(\xi_3 \mathbf{z}),
\end{equation}
where $\mathbf{z}$ is the $n_{\mathrm{S}}\times 1$ new design variable vector that would replace $\mathbf{b}$ in the optimization problem and $\boldsymbol \xi=[\xi_1, \xi_2, \xi_3]^{\mathrm{T}}$ is a vector with real constants as its elements.
Using~\eqref{Equ:Trans1}, the spectrum allocation problem for the special case system can be transformed into an equivalent standard convex problem given by
\begin{subequations} \label{P2V3}
\begin{alignat}{2}
 \mathbf {\hat{P}^{o}_{SC}:}\quad  &  \underset{\substack{\mathbf{p}, \mathbf{z}}}{\textrm{min}}
 ~~\quad -\mathbf{1}^{\textrm{T}}_{n_{\mathrm{S}}} \log (\mathbf{r})   \\
&  \textup{s. t.}
  \quad\quad\eqref{OptProbOrgB}, \eqref{OptProbOrgC},    \notag          \\
&   \quad \quad \quad \prod_{s=1}^{n_{\mathrm{S}}}z_{s}^{\xi_2}- z_{\textrm{tot}}\leqslant 0,      \label{OptProb2Sol2-C}  \\
&  \quad \quad \quad z_{\textrm{min}} \preceq \textbf{z} \preceq z_{\textrm{max}} ,   \label{OptProb2Sol2-A}
\end{alignat}
\end{subequations}
where  $z_{\textrm{tot}}=({\xi_3}^{-\xi_2}e^{{b}_{\textrm{tot}}-\xi_1 n_{\mathrm{S}}})^{n_{\mathrm{S}}}$, $z_{\textrm{min}}=\xi_3^{-1}e^{\frac{-\xi_1}{\xi_2}}$,  and $ z_{\textrm{max}}=\xi_3^{-1}e^{\frac{b_{\textrm{max}}-\xi_1}{\xi_2}}$. The transformed problem for the special case system, $\mathbf {\hat{P}^{o}_{SC}}$, can be solved efficiently by using standard convex problem solvers \cite{ConvexBoyedBook}.
The details of the convex optimization-based approach can be found in our previous work~\cite{akram2021TCOM,akram2022TVT}.

\section{Numerical Results}\label{Sec:Num}

The numerical results are obtained by considering a rectangular indoor environment of size $25~\textrm{m}\times 25~\textrm{m}$ where an AP, which is located at the center of the ceiling of the indoor environment, serves 15 users. We consider the spectrum with a bandwidth of 50 GHz that exist in either (i) the NACSR $sr_{n1}$ between $0.557$ and $0.671~\textrm{THz}$ or (ii) the NACSR $sr_{n2}$ between $0.752$ and $0.868~\textrm{THz}$ is used to serve the users.
We clarify that on one hand, the molecular absorption coefficient, $k(f)$, in $sr_{n2}$ can be modeled as an exponential function of frequency with minimal approximation errors. Hence, the spectrum allocation problem in $sr_{n2}$ is a special case of the general spectrum allocation problem considered in this work.
On the other hand, $k(f)$ in $sr_{n1}$ cannot be modeled as an exponential function of frequency. 

We implement the unsupervised learning model presented in Section \ref{Sec:UL_Solution} in Python. We utilize a five-layer DNN for the unsupervised learning model, which has 100, 100, 50, 25, and 30 neurons in the first, second, third, fourth, and fifth layers, respectively. The initial weights of the DNN are Gaussian random variables with zero mean and unit variance, and the initial biases are set to 0. The initial values of $\boldsymbol \lambda$ are set to a small constant of $0.1$.
The values of other system parameters, hyper-parameters of the unsupervised learning model, and the parameters of the convex optimization-based approach used for numerical results are summarized in Table \ref{tab1}.

\begin{table}[t]
\caption{Value of Parameters Used in Section~\ref{Sec:Num}.}
\begin{center}
\begin{tabular}{|l|l|}
\hline
\textbf{Parameter} & \textbf{Value}  \\
\hline
\multicolumn{2}{|c|}{\textbf{System Parameters}}  \\
\hline
Difference between the heights of AP and$\!\!\!\!$   & $1.7~\textrm{m}$ \\
 users & \\ \hline
Antenna gains, $G_{\textrm{A}}$, $G_{\textrm{U}}$  &   $30~\textrm{dBi}$, $20~\textrm{dBi}$   \\ \hline
 Noise power density, $N_{0}$  & $-174~\textrm{dBm/Hz}$   \\
\hline
Power budget, $p_{\textrm{tot}}$ & -5~$\textrm{dBm}$    \\ \hline
Upper bound on power per sub-band, $p_{\textrm{max}}\!\!\!\!$   & $\frac{5}{4}\frac{p_{\textrm{tot}}}{n_{\mathrm{I}}}$         \\ \hline
Upper bound on sub-band bandwidth, $b_{\textrm{max}}\!\!\!\!$  & $5~\textrm{GHz}$ \\ \hline
\multicolumn{2}{|c|}{\textbf{Hyper-Parameters of the Unsupervised Learning Model}}  \\
\hline
Learning rates, $\delta_{ \Theta}$ $\delta_{ \lambda}$   & 0.05, 0.025 \\
\hline
Number of iterations  &   500   \\ \hline
Total number of realization of $\textbf{d}$, $n_{\mathrm{T}}$  & $300$   \\
\hline
Activation function of 1st to 4th layer, $\sigma^{(\ell)}\!(~\!\!\!\!\cdot~\!\!\!\!)\!\!\!\!$ & ReLU    \\ \hline
Activation function of the 5th layer, $\sigma^{(\ell)}\!(~\!\!\!\!\cdot~\!\!\!\!)\!\!\!$ & $\!\!\vartheta \cdot$sigmoid, $\vartheta\!\!\in\!\! \{p_{\textrm{max}},\!b_{\textrm{max}}\}\!\!\!\!$     \\ \hline
\multicolumn{2}{|c|}{\textbf{Parameters used for the Convex Optimization-based Approach}}  \\
\hline
Exponential model parameters in $sr_{n1}$, $\boldsymbol \eta\!\!$  & $10^{1.83}\!$,$\scalebox{1}{-}10^{\scalebox{1}{-}10.04}\!$,$10^{\scalebox{1}{-}1.23}\!\!\!$ \\ \hline
Exponential model parameters in $sr_{n2}$, $\boldsymbol \eta\!\!$  & $10^{0.89}\!$,$\scalebox{1}{-}10^{\scalebox{1}{-}10.8}\!$,$\scalebox{1}{-}10^{\scalebox{1}{-}1.53}\!\!\!\!$ \\ \hline
Transformation model parameters, $\boldsymbol \xi$  &   $10^{9.7}$,$10^{10.7}$,$10^{-3}$   \\ \hline
\end{tabular}\label{tab1}
\end{center}\vspace{-4mm}
\end{table}

\begin{figure}[!t]
\centering\subfloat[\label{1a}]{ \includegraphics[width=0.45\columnwidth]{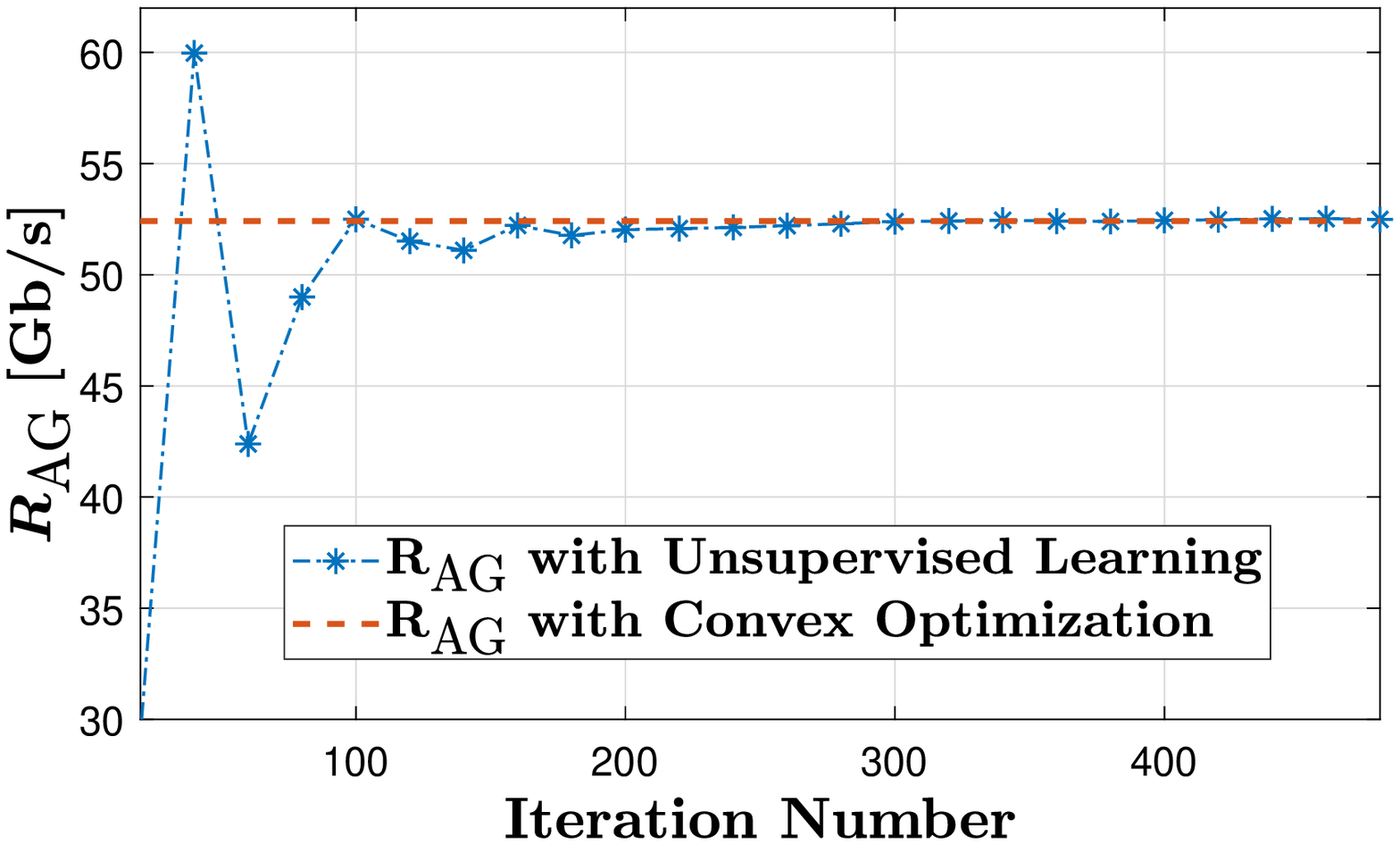}} \hspace{2 mm} \subfloat[\label{1b}]{\includegraphics[width=0.45\columnwidth]{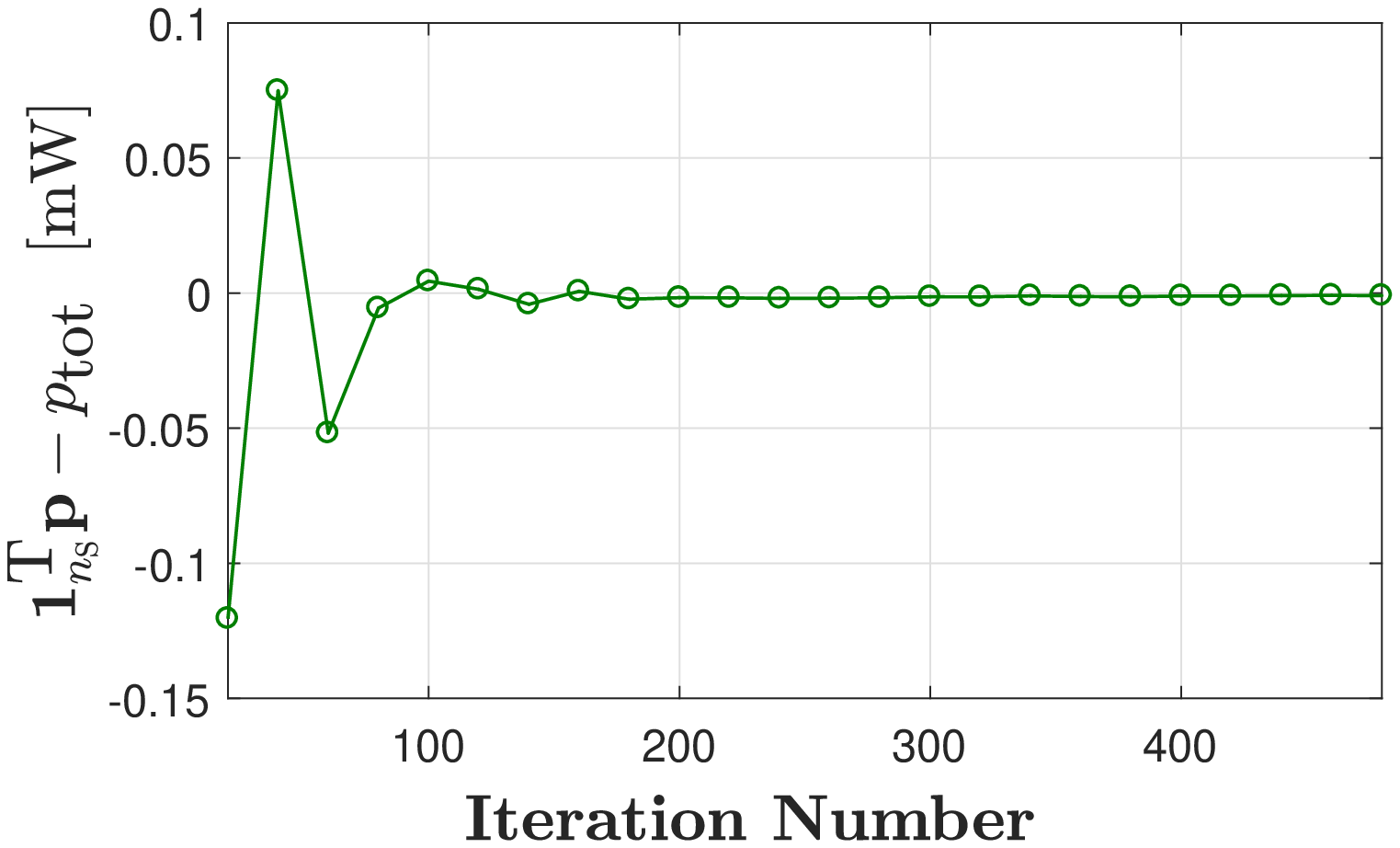}} \hspace{2 mm}
\subfloat[\label{1c} ]{\includegraphics[width=0.45\columnwidth]{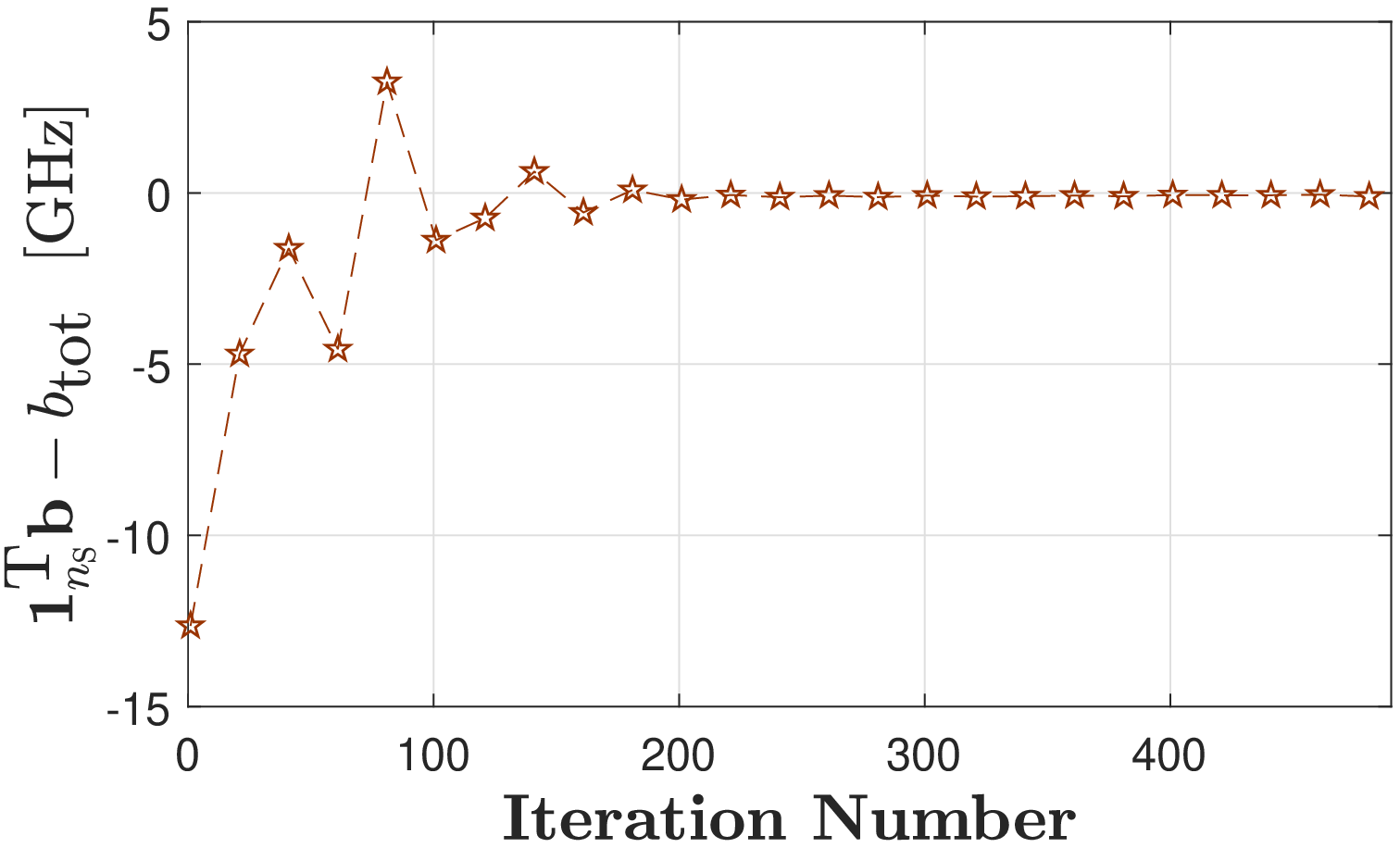}}
  \caption{The aggregated multiuser data rate, $R_{\textrm{AG}}$, and total power and bandwidth constraints satisfaction when $k(f)$ within the spectrum of interest can be modeled as an exponential function of frequency, i.e., in the special case system investigated in Section \ref{Sec:Special_Case}.}\label{Fig:NumFigA}
\end{figure}

\begin{figure}[!t]
\centering\subfloat[\label{2a}]{ \includegraphics[width=0.45\columnwidth]{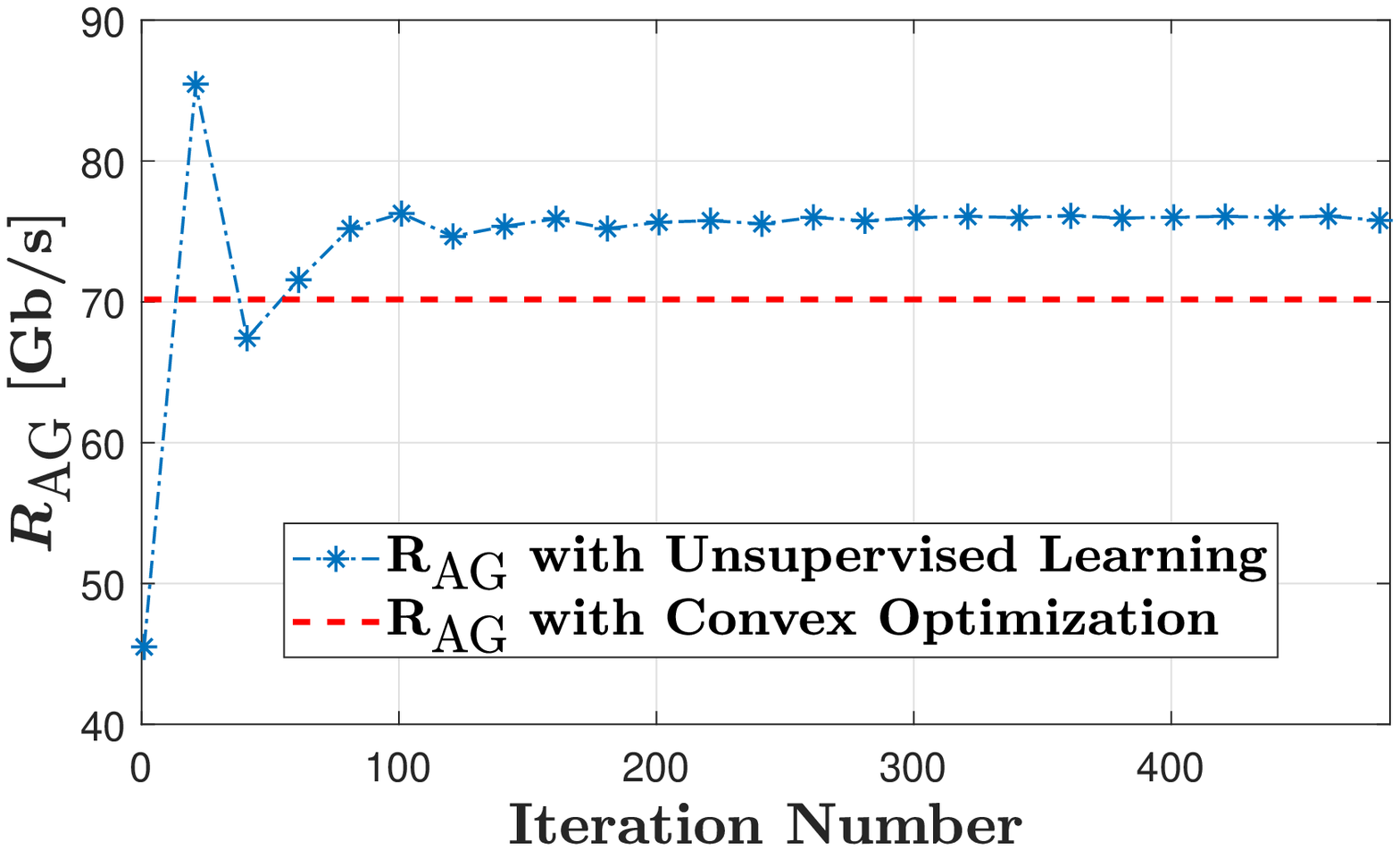}} \hspace{2 mm} \subfloat[\label{2b}]{\includegraphics[width=0.45\columnwidth]{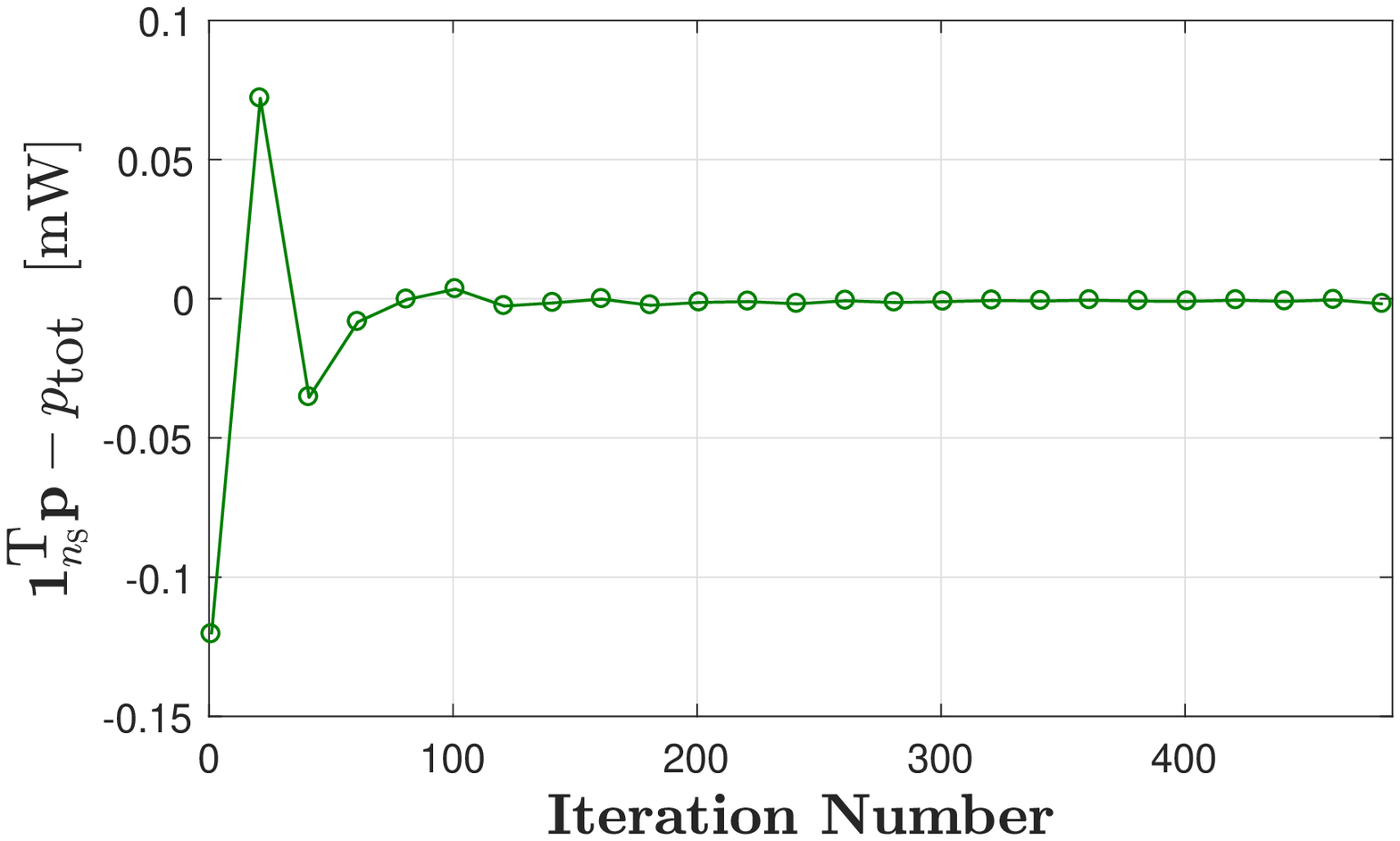}} \hspace{2 mm}
\subfloat[
\label{2c} ]{\includegraphics[width=0.45\columnwidth]{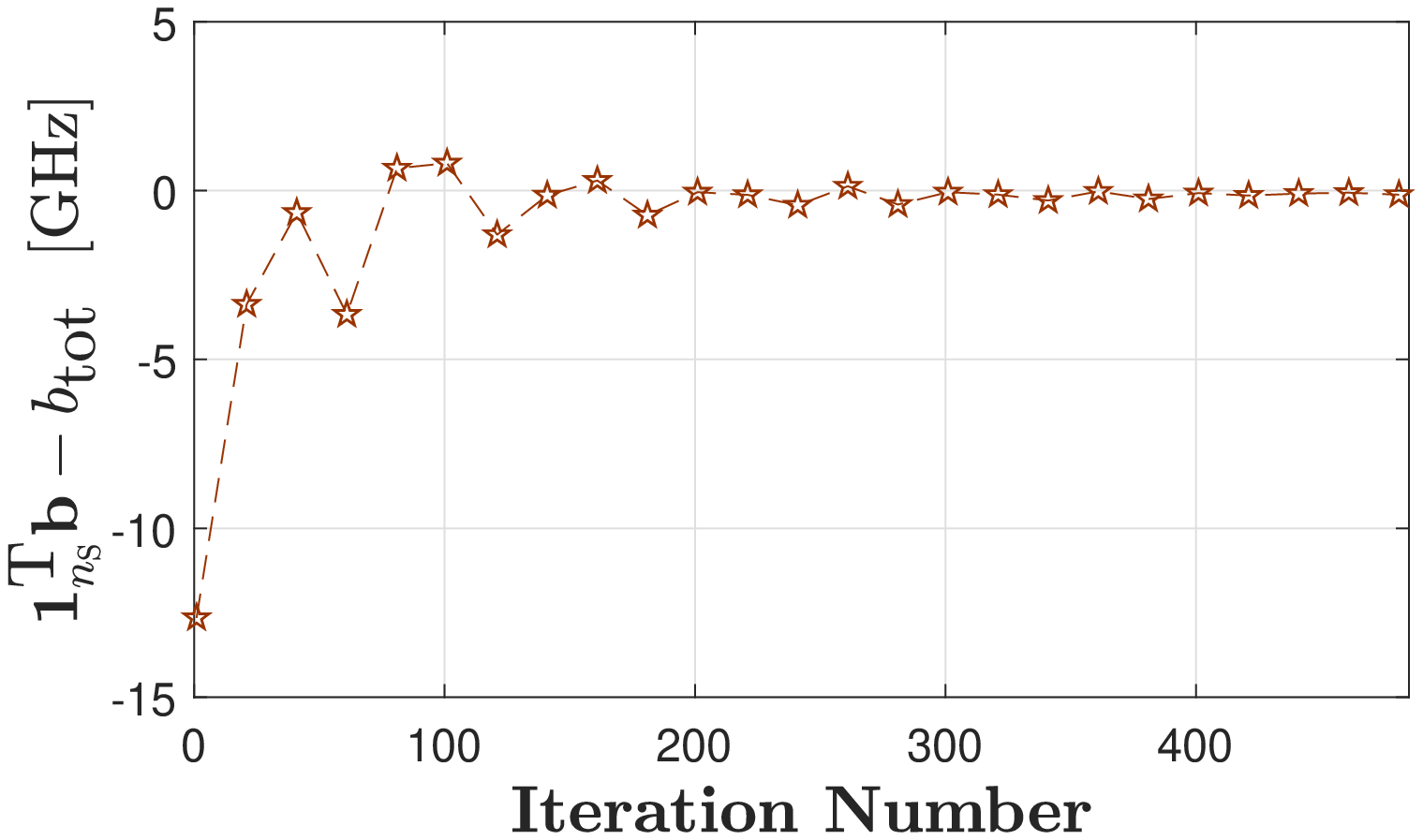}}
  \caption{The aggregated multiuser data rate, $R_{\textrm{AG}}$, and total power and bandwidth constraints satisfaction when $k(f)$ within the spectrum of interest cannot be modeled as an exponential function of frequency, i.e., in the generalized system investigated in Sections \ref{Sec:System} and \ref{Sec:UL_Solution}.}\label{Fig:NumFigB}
\end{figure}


In Fig.\ref{Fig:NumFigA}, we show the convergence of the proposed unsupervised learning-based approach
when $k(f)$ within the spectrum of interest can be modeled as an exponential function of frequency, i.e., considering the special case system investigated in Section \ref{Sec:Special_Case}. To this end, we plot the aggregated multiuser data rate, $R_{\textrm{AG}}=\mathbf{1}_{n_{\mathrm{S}}}^{\textrm{T}}\mathbf{r}$, the value of $\mathbf{1}_{n_{\mathrm{S}}}^{\textrm{T}}\mathbf{p}- p_{\textrm{tot}}$, and the value of $\mathbf{1}_{n_{\mathrm{S}}}^{\textrm{T}}\mathbf{b} - b_{\textrm{tot}}$ obtained from the unsupervised learning-based approach in Figs. \ref{Fig:NumFigA}(a), \ref{Fig:NumFigA}(b), and \ref{Fig:NumFigA}(c), respectively, when the spectrum allocation problem is implemented in $sr_{n2}$. Also, the value of $R_{\textrm{AG}}$ obtained from the convex optimization-based approach is plotted in Fig.\ref{Fig:NumFigA}(a).
We first observe from Fig.\ref{Fig:NumFigA}(a) that for several initial iterations, the value of $R_{\textrm{AG}}$ obtained using the unsupervised learning-based approach is higher than that obtained using the convex optimization-based approach. This is due to the occasional violation of the constraints \eqref{OptProbOrgB} and \eqref{OptProbOrgD} that occur at corresponding initial iterations, which can be validated by observing $\mathbf{1}_{n_{\mathrm{S}}}^{\textrm{T}}\mathbf{p}- p_{\textrm{tot}}$ and $\mathbf{1}_{n_{\mathrm{S}}}^{\textrm{T}}\mathbf{b} - b_{\textrm{tot}}$ in Figs. \ref{Fig:NumFigA}(b) and \ref{Fig:NumFigA}(c), respectively. Moreover, after 200 iterations, we observe that  the value of $R_{\textrm{AG}}$ obtained using the unsupervised learning-based approach converges to the optimal value obtained using the convex optimization-based approach, which shows the correctness of our proposed unsupervised learning-based approach.
Finally, we observe from Figs. \ref{Fig:NumFigA}(b) and \ref{Fig:NumFigA}(c) that after 200 iterations, the values of $\mathbf{1}_{n_{\mathrm{S}}}^{\textrm{T}}\mathbf{p}- p_{\textrm{tot}}$ and $\mathbf{1}_{n_{\mathrm{S}}}^{\textrm{T}}\mathbf{b} - b_{\textrm{tot}}$ of the proposed unsupervised learning-based approach are very close to zero, which reflects that constraints \eqref{OptProbOrgB} and \eqref{OptProbOrgD} are satisfied when the proposed unsupervised learning-based approach is used.

\begin{figure}[!t]
\centering
\includegraphics[width=0.7\columnwidth]{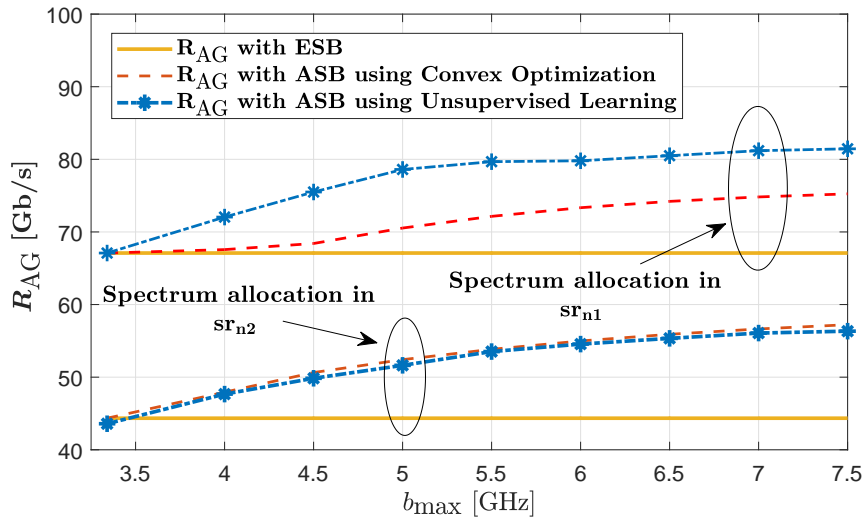}
\vspace{-3mm}
\caption{Aggregated multiuser data rate, $R_{\textrm{AG}}$, versus the upper bound on sub-band bandwidth, $b_{\textrm{max}}$, when spectrum allocation in $sr_{n1}$ and $sr_{n2}$.}\label{Fig:NumFigC} \vspace{-3mm}
\end{figure}

We next show the convergence of the proposed unsupervised learning-based approach when $k(f)$ within the spectrum of interest cannot be modeled as an exponential function of frequency, i.e., considering the generalized system investigated in Sections \ref{Sec:System} and \ref{Sec:UL_Solution}. To this end, we plot $R_{\textrm{AG}}$, $\mathbf{1}_{n_{\mathrm{S}}}^{\textrm{T}}\mathbf{p}- p_{\textrm{tot}}$, and $\mathbf{1}_{n_{\mathrm{S}}}^{\textrm{T}}\mathbf{b} - b_{\textrm{tot}}$ in Figs. \ref{Fig:NumFigB}(a), \ref{Fig:NumFigB}(b), and \ref{Fig:NumFigC}(c), respectively, when the spectrum allocation problem is implemented in $sr_{n1}$. We clarify that  in $sr_{n1}$, $k(f)$ cannot be modeled as an exponential function of frequency with minimal approximation errors. However, for the sake of comparison,  we obtain approximate solutions to the spectrum allocation problem in $sr_{n1}$ using the convex optimization approach while utilizing an inaccurate approximation for $k(f)$ in $sr_{n1}$, and plot the resulting $R_{\textrm{AG}}$ in Fig.\ref{Fig:NumFigB}(a). Once again, we observe the overshoot in $R_{\textrm{AG}}$, $\mathbf{1}_{n_{\mathrm{S}}}^{\textrm{T}}\mathbf{p}- p_{\textrm{tot}}$, and $\mathbf{1}_{n_{\mathrm{S}}}^{\textrm{T}}\mathbf{b} - b_{\textrm{tot}}$ for several initial iterations and the satisfaction of the constraints \eqref{OptProbOrgB} and \eqref{OptProbOrgD} after 200 iterations in Figs. \ref{Fig:NumFigB}(a)-\ref{Fig:NumFigB}(c), which are similar to Figs. \ref{Fig:NumFigA}(a)-\ref{Fig:NumFigA}(c). Apart from them, we further observe that the value of $R_{\textrm{AG}}$ obtained using the unsupervised learning-based approach converges to a value higher than that obtained using the convex optimization-based approach. This is because when the convex optimization-based approach is used, where an exponential function with high approximation errors is used to model $k(f)$, only the sub-optimal $R_{\textrm{AG}}$ is obtained. This shows the significance of our proposed unsupervised learning-based approach, i.e., it gives much higher $R_{\textrm{AG}}$ for the spectrum allocation problem when $k(f)$ within the spectrum of interest cannot be modeled as an exponential function of frequency.

Finally, we plot $R_{\textrm{AG}}$ achieved by the spectrum allocation strategy with ESB \cite{HBM2,2017N5,HBM3,2019_Chong_DABM2} and the spectrum allocation strategy with ASB that is obtained from both convex optimization and unsupervised learning-based approaches, versus the upper bound on sub-band bandwidth, $b_{\textrm{max}}$, in Fig.\ref{Fig:NumFigC}. We first observe that the proposed spectrum strategy with ASB achieves a significantly higher $R_{\textrm{AG}}$ compared to the strategy with ESB for different $b_{\textrm{max}}$, which demonstrates the benefits of our proposed strategy with ASB. Second, we observe that in $sr_{n1}$, the value of $R_{\textrm{AG}}$ obtained using the unsupervised learning-based approach converges to values higher than that obtained using the convex optimization-based approach for all $b_{\textrm{max}}$. This again shows the significance of our proposed  unsupervised learning-based approach to obtain higher $R_{\textrm{AG}}$, especially when $k(f)$ within the spectrum of interest cannot be modeled as an exponential function of frequency.

\section{Conclusions}

We investigated multi-band-based spectrum allocation with ASB for multiuser THzCom systems. We first formulated an optimization problem and then proposed an unsupervised learning-based approach to obtaining near-optimal sub-band bandwidth and transmit power. In the proposed unsupervised learning-based approach,  we first employed an offline training phase to train a DNN while utilizing a loss function inspired by the Lagrangian of the formulated problem. Then using the trained DNN, we approximated the near-optimal solutions to the optimization problem for the given distance vector. Numerical results showed that when the values of the molecular absorption coefficient within the spectrum cannot be modeled as an exponential function of frequency, the data rate obtained by our proposed unsupervised learning-based approach outperforms that obtained by the existing convex optimization-based approach.

\bibliographystyle{IEEEtran}

\begin{thebibliography}{10}
\providecommand{\url}[1]{#1}
\csname url@samestyle\endcsname
\providecommand{\newblock}{\relax}
\providecommand{\bibinfo}[2]{#2}
\providecommand{\BIBentrySTDinterwordspacing}{\spaceskip=0pt\relax}
\providecommand{\BIBentryALTinterwordstretchfactor}{4}
\providecommand{\BIBentryALTinterwordspacing}{\spaceskip=\fontdimen2\font plus
\BIBentryALTinterwordstretchfactor\fontdimen3\font minus
  \fontdimen4\font\relax}
\providecommand{\BIBforeignlanguage}[2]{{%
\expandafter\ifx\csname l@#1\endcsname\relax
\typeout{** WARNING: IEEEtran.bst: No hyphenation pattern has been}%
\typeout{** loaded for the language `#1'. Using the pattern for}%
\typeout{** the default language instead.}%
\else
\language=\csname l@#1\endcsname
\fi
#2}}
\providecommand{\BIBdecl}{\relax}
\BIBdecl

\bibitem{2020_Mag6G_Marco_UseCasesandTechnologies}
M.~{Giordani}, M.~{Polese}, M.~{Mezzavilla}, S.~{Rangan}, and M.~{Zorzi},
  ``Toward {6G} networks: Use cases and technologies,'' \emph{IEEE Commun.
  Mag.}, vol.~58, no.~3, pp. 55--61, Mar. 2020.

\bibitem{akram2022IEEENetwork}
\BIBentryALTinterwordspacing
A.~Shafie, N.~Yang, C.~Han, J.~M. Jornet, M.~Juntti, and T.~K\"urner,
  ``Terahertz communications for {6G} and beyond wireless networks: Challenges,
  key advancements, and opportunities,'' July 2022. [Online]. Available:
  \url{https://arxiv.org/abs/2207.11021}
\BIBentrySTDinterwordspacing

\bibitem{2022_Sarieddeen_ComMag}
H.~Sarieddeen, N.~Saeed, T.~Y. Al-Naffouri, and M.-S. Alouini, ``Next
  generation terahertz communications: A rendezvous of sensing, imaging, and
  localization,'' \emph{IEEE Commun. Mag.}, vol.~58, no.~5, pp. 69--75, 2020.

\bibitem{2020_WCM_THzMag_TerahertzNetworks}
M.~{Polese}, J.~M. {Jornet}, T.~{Melodia}, and M.~{Zorzi}, ``Toward end-to-end,
  full-stack {6G} terahertz networks,'' \emph{IEEE Commun. Mag.}, vol.~58,
  no.~11, pp. 48--54, Nov. 2020.

\bibitem{2020_WCM_THzMag_Standardization}
V.~{Petrov}, T.~{Kurner}, and I.~{Hosako}, ``{IEEE} 802.15.3d: First
  standardization efforts for sub-terahertz band communications toward 6{G},''
  \emph{IEEE Commun. Mag.}, vol.~58, no.~11, pp. 28--33, Nov. 2020.

\bibitem{HBM2}
C.~{Han} and I.~F. {Akyildiz}, ``Distance-aware bandwidth-adaptive resource
  allocation for wireless systems in the terahertz band,'' \emph{IEEE Trans.
  THz Sci. Technol.}, vol.~6, no.~4, pp. 541--553, July 2016.

\bibitem{2017N5}
A.~{Moldovan}, P.~{Karunakaran}, I.~F. {Akyildiz}, and W.~H. {Gerstacker},
  ``Coverage and achievable rate analysis for indoor terahertz wireless
  networks,'' in \emph{Proc. IEEE Int. Conf. Commun. (ICC)}, Paris, France, May
  2017, pp. 1--7.

\bibitem{HBM3}
C.~{Han}, A.~O. {Bicen}, and I.~F. {Akyildiz}, ``Multi-wideband waveform design
  for distance-adaptive wireless communications in the terahertz band,''
  \emph{IEEE Trans. Signal Process.}, vol.~64, no.~4, pp. 910--922, Feb. 2016.

\bibitem{2019_Chong_DABM2}
X.~{Zhang}, C.~{Han}, and X.~{Wang}, ``Joint beamforming-power-bandwidth
  allocation in terahertz {NOMA} networks,'' in \emph{Proc. Int. Conf. Sensing,
  Commun., Netw. (SECON)}, Boston, MA, USA, Sept. 2019, pp. 1--9.

\bibitem{akram2021TCOM}
A.~Shafie, N.~Yang, S.~A. Alvi, C.~Han, S.~Durrani, and J.~M. Jornet,
  ``Spectrum allocation with adaptive sub-band bandwidth for terahertz
  communication systems,'' \emph{IEEE Trans. Commun.}, vol.~70, no.~2, pp.
  1407--1422, Jan. 2022.

\bibitem{akram2022TVT}
\BIBentryALTinterwordspacing
A.~Shafie, N.~Yang, C.~Han, and J.~M. Jornet, ``Novel spectrum allocation among
  multiple transmission windows for terahertz communication systems,'' July
  2022. [Online]. Available: \url{https://arxiv.org/abs/2207.02401}
\BIBentrySTDinterwordspacing

\bibitem{2011_Jornet_TWC}
J.~M. {Jornet} and I.~F. {Akyildiz}, ``Channel modeling and capacity analysis
  for electromagnetic wireless nanonetworks in the terahertz band,'' \emph{IEEE
  Trans. Wireless Commun.}, vol.~10, no.~10, pp. 3211--3221, Oct. 2011.

\bibitem{akram2020JSAC}
A.~Shafie, N.~Yang, S.~Durrani, X.~Zhou, C.~Han, and M.~Juntti, ``Coverage
  analysis for 3{D} terahertz communication systems,'' \emph{IEEE J. Sel. Areas
  Commun.}, vol.~39, no.~6, pp. 1817--1832, June 2021.

\bibitem{2020_Chong_TWC_DistanceAdaptiveAbsorptionPeakModulation}
W.~{Gao}, Y.~{Chen}, C.~{Han}, and Z.~{Chen}, ``Distance-adaptive absorption
  peak modulation for terahertz covert communications,'' \emph{IEEE Trans.
  Wireless Commun.}, vol.~20, no.~3, pp. 2064--2077, Nov. 2020.

\bibitem{ConvexBoyedBook}
S.~Boyd and L.~Vandenberghe, \emph{Convex Optimization}.\hskip 1em plus 0.5em
  minus 0.4em\relax Cambridge, U.K.: Cambridge Univ. Press, 2004.

\bibitem{2019_TSP_DNN_Mark}
M.~Eisen, C.~Zhang, L.~F.~O. Chamon, D.~D. Lee, and A.~Ribeiro, ``Learning
  optimal resource allocations in wireless systems,'' \emph{IEEE Trans. Signal
  Process.}, vol.~67, no.~10, pp. 2775--2790, May 2019.

\bibitem{2022_Chunhui_MLforSecurity}
C.~Li, C.~She, N.~Yang, and T.~Q.~S. Quek, ``Secure transmission rate of short
  packets with queueing delay requirement,'' \emph{IEEE Trans. Wireless
  Commun.}, vol.~21, no.~1, pp. 203--218, Jan. 2022.

\end{thebibliography}


\end{document}